\documentclass[10pt]{article}

\usepackage{scicite}

\usepackage{times}

\usepackage{amsmath}
\usepackage{graphicx}
\usepackage{url}
\usepackage{booktabs}
\usepackage{multirow}
\usepackage{caption}
\usepackage{subcaption}
\usepackage{xcolor}
\usepackage{wasysym}
\usepackage{soul}

\usepackage{placeins}

\newcommand{\ci}[2]{\textcolor{gray}{\tiny \ [#1,#2]}}

\newcommand{\cg}[1]{\textcolor{gray}{#1}}

\newenvironment{sciabstract}{%
\begin{quote} \bf}
{\end{quote}}

\title{Advancing human-centric AI for robust X-ray analysis \\through holistic self-supervised learning}

\author
{Theo Moutakanni$^{1,2\ast}$, Piotr Bojanowski$^{1}$, Guillaume Chassagnon$^{3,4}$,  Céline Hudelot$^{2}$,\\
Armand Joulin$^{1}\dagger$, Yann LeCun$^{1}$, Matthew Muckley$^{1}$, Maxime Oquab$^{1}$, Marie-Pierre Revel$^{3,4}$,\\
Maria Vakalopoulou$^{2}$\\
\\
\normalsize{$^{1}$FAIR at Meta}\\
\normalsize{$^{2}$MICS, CentraleSupélec, Université Paris-Saclay, France}\\
\normalsize{$^{3}$Department of Radiology, Hôpital Cochin, AP-HP.Centre Université Paris Cité, France}\\
\normalsize{$^{4}$Université Paris Cité, France}
\\
\normalsize{$^\ast$To whom correspondence should be addressed; E-mail:  theomoutakanni@meta.com}\\
\normalsize{$\dagger$ Work done while at Meta}
}

\date{}

\renewcommand{\figurename}{Fig.}

\begin{document}

\maketitle

\begin{sciabstract}
AI Foundation models are gaining traction in various applications, including medical fields like radiology. However, medical foundation models are often tested on limited tasks, leaving their generalisability and biases unexplored. 
We present RayDINO, a large visual encoder trained by self-supervision on 873k chest X-rays.  
We compare RayDINO to previous state-of-the-art models across nine radiology tasks, from classification and dense segmentation to text generation, and provide an in-depth analysis of population, age and sex biases of our model. 
Our findings suggest that self-supervision allows patient-centric AI proving useful in clinical workflows and interpreting X-rays holistically. 
With RayDINO and small task-specific adapters, we reach state-of-the-art results and improve generalization to unseen populations while mitigating bias, illustrating the true promise of foundation models: versatility and robustness.
\end{sciabstract}

\section*{Introduction}
\label{sec:intro}

\begin{figure}[!htb]
  \centering
  \includegraphics{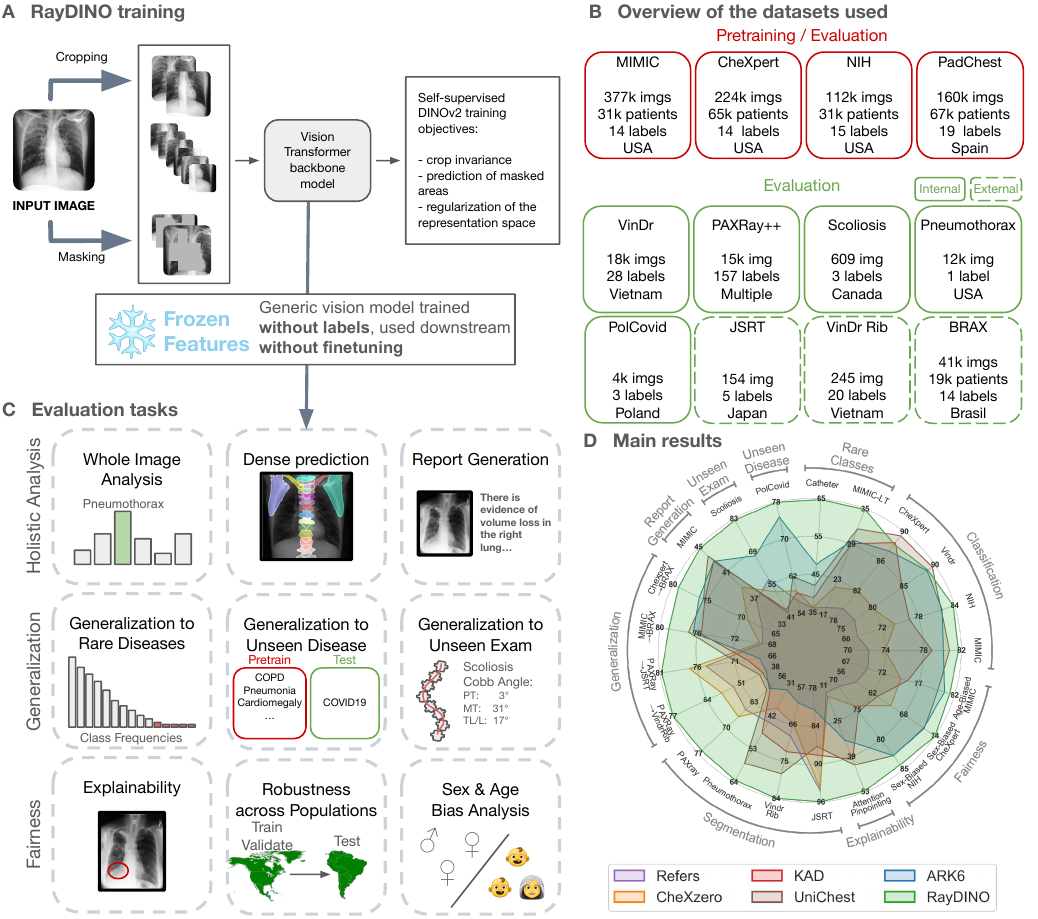}
  \caption{
\textbf{We introduce RayDINO, a foundation model for holistic and fairer analysis of chest X-rays.}
\textbf{a)} RayDINO is a 307M parameter vision transformer trained using the DINOv2 self-supervised objectives and applied as-is to all downstream tasks without any modification or specialization of its parameters.
\textbf{b)} RayDINO is trained on four different datasets from the USA and Europe, comprising over 870k X-rays, and is tested on eleven datasets from seven countries across four continents in both internal and external settings.
\textbf{c)} RayDINO is evaluated on tasks divided into three categories.
\textbf{d)} RayDINO significantly outperforms all other models on 21 benchmarks and consistently delivers excellent performance (AUROC for Classification and Fairness, macro Accuracy for Explainability, mDice for Segmentation, AUROC for BRAX Generalization, and mDice for PAXRay Generalization, CheXbert vector similarity for Report Generation, Pearson's correlation for Unseen Exam, macro Accuracy for Unseen Disease, AUPRC for Rare Classes). All details about the method and implementation are available in the Methods section.
  }
  \label{fig:fig1}
\end{figure}

Clinical workflows require a holistic analysis and interpretation of patient data~\cite{Bilalić2022}. 
While humans excel at contextualizing information, current machine learning models often fall short, extracting only the minimal information necessary to solve a training task defined by supervision, without considering the broader context. 
This is particularly evident in radiology, where chest X-rays, one of the most common and easily reachable types of exam to diagnose a variety of conditions affecting the lung and adjacent organs~\cite{rajpurkar2018deep}. However, providing separate AI models for each organ and anomaly visible in chest X-rays is impractical and can be unfeasible, especially in scenarios where access to annotated data is limited.

Foundation models offer a solution to these challenges. They are trained in a more general manner, thereby capturing most available information in the data. Those models are becoming increasingly popular in medical tasks, with many recent papers reporting good performances~\cite{Azizi2023,Zhou2023,chen2024uni}.
Modern AI systems often rely heavily on natural language for their training \cite{radford2021learning}. These algorithms learn image features by embedding paired images and textual descriptions in a shared feature space. However, the structure and complexity of language used in medical reports differ significantly from image captions derived from web data and training such models for medical tasks can be complex due to the heterogeneous nature of the data. Recent studies have raised concerns about the quality of automatic text-based data labeling\cite{zhang2022improving} and the reliability of AI systems trained on reports from medical images\cite{ramesh2022improving}. 
Such reports often include references to past examinations of a patient, and therefore a given report does not always correspond to the information present in the image seen during model training; this misalignment is a central issue for medical vision models built using natural language supervision.
Moreover, country-specific or brittle guidelines~\cite{doi:10.1148/radiol.2021203742,10.1001/jama.2021.13304,doi:10.2214/AJR.17.18907}, shaping the text reports, can hinder the robustness of foundation models when applied on new populations.
In this study, we propose an approach that circumvents these annotation-related challenges by 
training a vision foundation model using self-supervised learning on a large dataset of $873$k images compiled from four public X-ray image datasets. Self-supervised learning, which does not require any associated textual data or human annotations, learns features by modeling correlations within the visual data itself, leading to more holistic~\cite{oquab2024dinov}, patient-centric and unbiased representations~\cite{liu2022selfsupervised,Doron2023.06.16.545359}.

We introduce RayDINO, the largest vision transformer (ViT)~\cite{dosovitskiy2020image} model of $307$M parameters that can generate holistic representations of chest X-rays using the self-supervised DINOv2 method~\cite{oquab2024dinov} (Fig.~\ref{fig:fig1}a). Our model is trained on four and evaluated on eleven publicly available datasets (Fig.~\ref{fig:fig1}b) across nine medical-imaging related tasks (Fig.~\ref{fig:fig1}c)  divided in twenty-one benchmarks comprising more than two hundred classes. RayDINO surpasses other task-specific supervised models (Fig.~\ref{fig:fig1}d) and achieves this by using the same frozen encoder on all downstream tasks while training only small task-specific adapters. 
To show the holistic and robust X-ray analysis of RayDINO, we compared our model using a wide variety of benchmark, ranging from classification, and dense segmentation of organs and diseases, to report generation. Our tasks notably include clinically rare diseases and other findings underrepresented in data. 
As part of our extensive quantitative evaluation, we analyzed the bias and potential limitations of such foundation models. Our findings indicate that our model's  representation provides state-of-the-art performance, which surpasses the performance of models specifically designed to address these tasks, while being better at handling worst minority groups and external demographics. Finally, besides its state-of-the-art performances, our model provides also impressive interpretability, validated by expert radiologists.

More specifically, RayDINO is trained on four datasets originating from the USA (MIMIC~\cite{johnson2019mimic}, CheXpert~\cite{irvin2019chexpert}, NIH~\cite{wang17nih}) and Spain (PadChest~\cite{Bustos_2020}). We evaluated its performance using eleven datasets from seven countries across four continents. These include MIMIC, NIH, CheXpert, and SIIM-ACR Pneumothorax~\cite{zawacki2019siimacr} from the USA; VinDr-CXR~\cite{nguyen2020vindrcxr} and Vindr-RibCXR ~\cite{nguyen2021vindrribcxr} from Vietnam; POLCOVID~\cite{suwalska2023polcovid} from Poland; BRAX~\cite{reis2022brax} from Brazil; JSRT~\cite{shiraishi2000development} from Japan; AASCE2019 Scoliosis~\cite{wu2017automatic} from Canada; and PAXRay++~\cite{Seibold_2022_BMVC,Seibold_2023_CXAS}, which consists of synthetic 2D projections of CT-scans primarily from the USA and China. We also enhance the granularity of annotations of MIMIC and NIH by incorporating new annotations from RANZCR CLiP Catheter~\cite{Seah2020RANZCRCLiP} and MIMIC Long-Tail~\cite{holste2023cxr}.

We benchmarked our model against five state-of-the-art supervised models: Refers~\cite{zhou2022generalized}, CheXzero~\cite{tiu2022expert}, 
KAD-512~\cite{zhang2023knowledge}, UniChest~\cite{dai2023unichest}, and ARK6~\cite{ma2023foundation}. Refers, CheXzero and KAD are trained on $377$k images from the MIMIC dataset using radiologists' free-text reports. UniChest is pretrained on $731$k X-rays using a combination of text reports and automatically-generated labels from the CheXpert, MIMIC, NIH, and Vindr datasets. ARK6, on the other hand, does not use text reports but is pretrained on the same datasets as UniChest using direct label annotations that match the classes from our evaluations. It's important to note that these models were fully fine-tuned end-to-end using in-domain labels or text, while our approach uses a frozen backbone pretrained without any human supervision. This establishes a challenging benchmark for RayDINO which has significantly fewer supervised parameters - between 100 and 400 times fewer compared to UniChest and ARK6.

Details about RayDINO along with metrics and experimental protocols are provided in the Materials and Methods section.

\section*{Results}

\begin{figure}[p]
  \centering
  \includegraphics{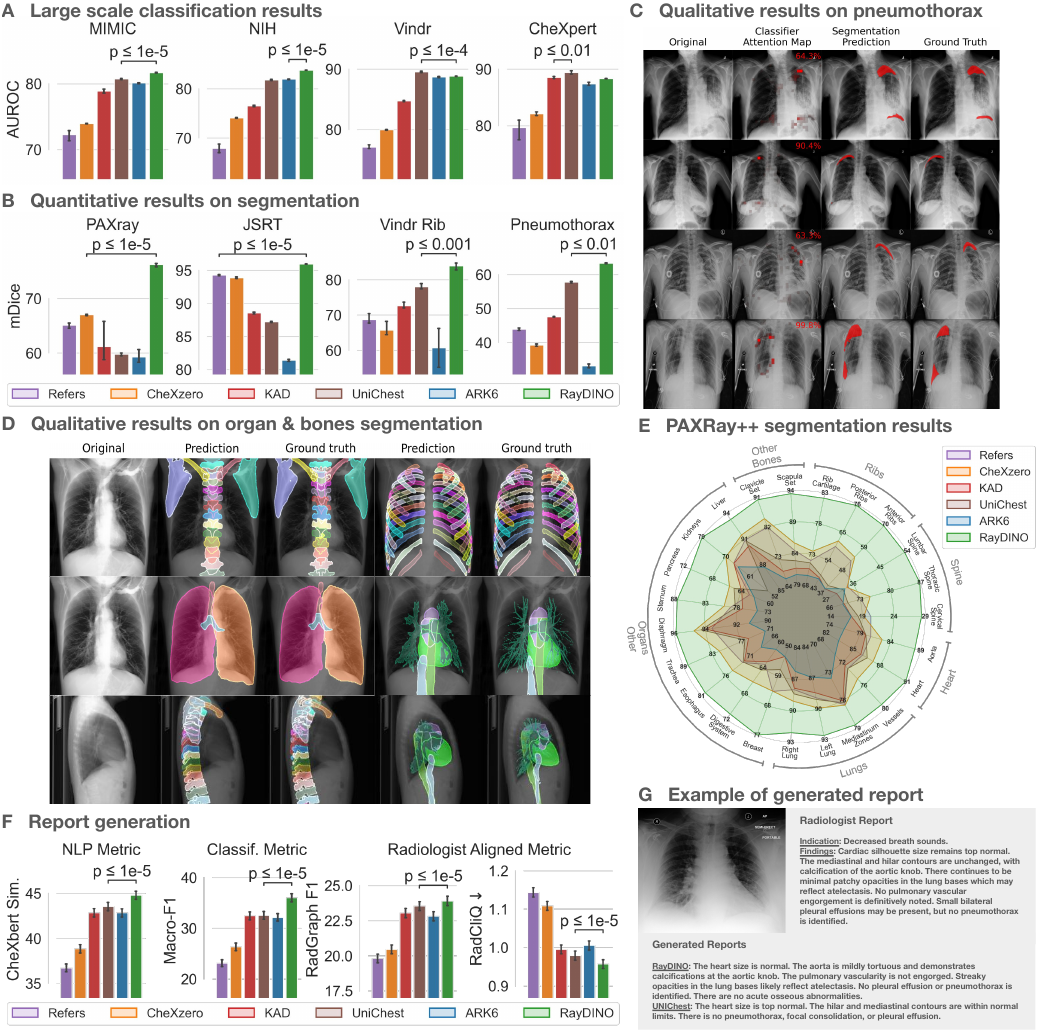}
  \caption{
    \textbf{RayDINO's holistic analysis of chest X-rays.}
    \textbf{a)} AUROC comparisons on 4 classification datasets from the USA and Vietnam including 38 different findings.
    \textbf{b)} mDice comparisons on 4 segmentation datasets from Japan, Vietnam, the USA, China and multiple other countries including 157 categories.
    \textbf{c)} Qualitative visualisation of RayDINO's prediction on four pneumothorax cases. \textbf{1\textsuperscript{st} column}: original image. \textbf{2\textsuperscript{nd} column}: interpretable attention maps showing where the classifier trained without pixel supervision is looking at. \textbf{3\textsuperscript{rd} column} segmentation prediction trained using pixel supervision. \textbf{4\textsuperscript{th} column}: radiologist ground-truth.
    \textbf{d)} Qualitative results for organ and bone segmentation on the PAXRay++ dataset.
    \textbf{e)} mDice segmentation comparisons macro-averaged per organs and bones.
    \textbf{f)} Report generation comparison including natural language processing metric, classification-based metric and radiologist-aligned metrics.
    \textbf{g)} A chest X-ray along with a radiologist report and two generated reports by RayDINO and UNIChest.
  }
  \label{fig:fig2}
\end{figure}

\paragraph{Unified Approach for Comprehensive Radiology Interpretation: Classification, Segmentation, and Report Generation.} 
To demonstrate the comprehensive analysis of X-ray images that our model, RayDINO, can perform, we tested it against a broad range of established benchmarks. These benchmarks include eight datasets with more than $200$ different classes to predict. They are categorized into three main tasks typically assessed in the literature. It's important to note that RayDINO's parameters are frozen and not updated during this process, making them fully unsupervised. Labels are only used to train a small task-specific adapter for each dataset and each benchmarked model. The results of these evaluations are detailed in the subsequent section and summarized in Fig.\ref{fig:fig2}, providing both qualitative and quantitative insights.

In our first set of experiments, we performed multilabel
classification on four datasets including a total of 38 different findings. Fig.~\ref{fig:fig2}a indicates that RayDINO significantly outperforms the best alternatives on the MIMIC dataset with $81.7$ AUROC (95\%CI [$81.6$-$81.8$], $+1.0$ over UniChest, $p\leq1\mathrm{e}{-5}$) and the NIH dataset with $83.8$ AUROC (95\%CI [$83.7$-$83.8$], $+1.9$ over ARK6, $p\leq1\mathrm{e}{-5}$). Furthermore, despite using $100$x less supervised parameters, RayDINO remains competitive on CheXpert with $88.3$ AUROC (95\%CI [$88.2$-$88.4$], $-1.1$ below UniChest, $p\leq1\mathrm{e}{-5}$) and on Vindr which is out-of-domain for RayDINO, contrary to ARK6 and UniChest, with $88.8$ AUROC (95\%CI [$88.8$-$88.8$], $-0.7$ below UniChest, $p\leq1\mathrm{e}{-5}$).

Then, we evaluated RayDINO on segmentation tasks on four datasets, three of which are on soft organs and bones segmentation to evaluate anatomical understanding while the other is on pneumothorax segmentation to show the precise localization capabilities of RayDINO's features. We report results in Fig.~\ref{fig:fig2}b and prove that it exhibits superior performance compared to all alternatives, on all benchmarks.
For bone and organ segmentation, our model outperforms all competitors with $96.0\%$ mDice on JSRT (95\%CI [$95.9$-$96.0$], $+1.7$ over Refers, $p\leq1\mathrm{e}{-5}$), $83.9\%$ mDice on Vindr Rib (95\%CI [$82.3$-$85.5$], $+5.9$ over UniChest, $p\leq1\mathrm{e}{-3}$), and $76.2\%$ mDice on the more challenging PAXRay++ dataset (95\%CI [$75.7$-$76.7$], $+11.1$ over CheXzero, $p\leq1\mathrm{e}{-5}$). We present PAXRay results macro-averaged per organ in Fig.~\ref{fig:fig2}e and demonstrate that RayDINO surpasses other models on all organs ($p\leq1\mathrm{e}{-5}$). RayDINO also performs significantly better at segmenting pneumothorax with $63.3\%$ mDice (95\%CI [$61.7$-$64.9$], $+5.6$ over UniChest, $p\leq0.01$).
We show the quality of RayDINO's predictions with the ground-truth segmentations on pneumothorax localization in Fig.~\ref{fig:fig2}c and on organ segmentation in Fig.~\ref{fig:fig2}d.

Finally, we challenged RayDINO for radiology report generation on MIMIC~\cite{johnson2019mimic} reports. This task requires a comprehensive understanding of images and is crucial with the advent of advanced systems able to interact with patients and radiologists through text.
As shown in Fig.\ref{fig:fig2}f, RayDINO outperforms UniChest and all other models on Natural Language Processing metrics ($44.8$ CheXbert similarity score, 95\%CI [$44.0$-$45.6$], $+1.2$, $p\leq1\mathrm{e}{-5}$), on classification metrics ($36.1$ macro F1, 95\%CI [$34.4$-$37.6$], $+3.4$, $p\leq1\mathrm{e}{-5}$) and on radiologist aligned metrics ($23.9$ RadGraph-F1, 95\%CI [$23.3$-$24.6$], $+0.4$, $p\leq1\mathrm{e}{-5}$ and $3.07$ RadCliQ, lower is better, 95\%CI [$3.04$-$3.11$], $-0.03$, $p\leq1\mathrm{e}{-5}$). We show in supplementary table~\ref{tab:report} that RayDINO even outperformed the best specialised model MAIRA-1~\cite{hyland2023maira} on radiologist aligned metrics (CheXbert similarity, RadCliQ and RadGraph-F1).
An example of a generated report is presented in Fig.~\ref{fig:fig2}g.

\paragraph{Harnessing Self-Supervised Learning for the Identification of Rare or Under-Represented Classes.}
We tested RayDINO more thoroughly by evaluating its performance in real-life medical and clinical situations. We focused on its ability to deal with rare health conditions that cannot be annotated on thousands of images. This assessment is crucial to determine the practical utility of our system in routine clinical practice and its potential for large-scale deployment.

\begin{figure}[!htb]
  \centering
  \includegraphics{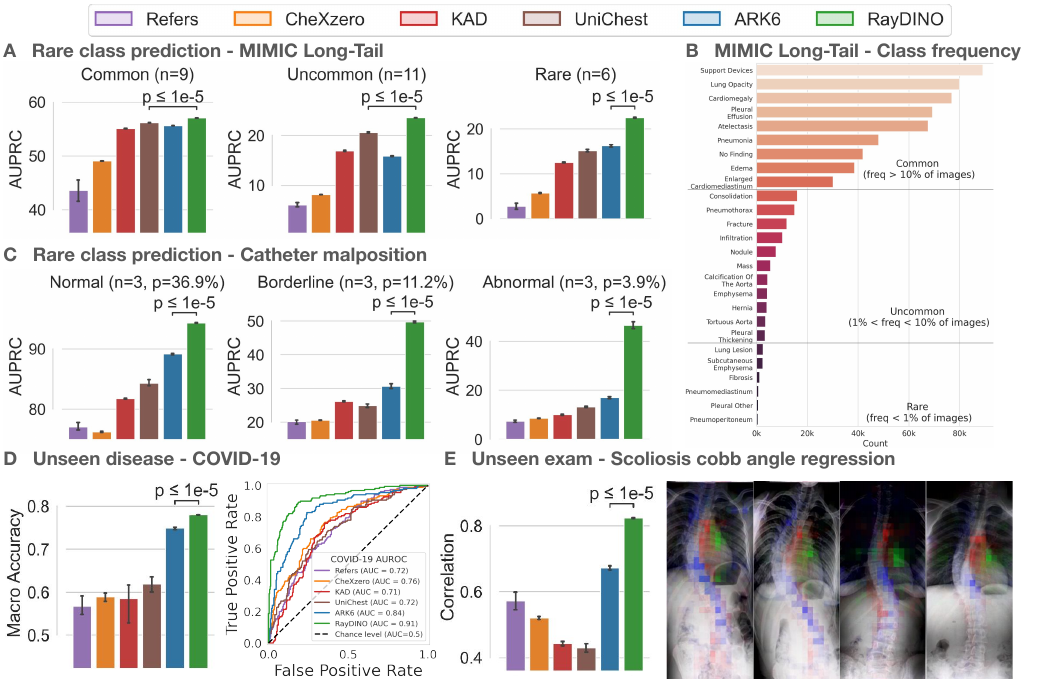}
  \caption{
    \textbf{Generalization evaluation on rare or unseen diseases and on unseen exam.}
    \textbf{a)} AUPRC comparisons of RayDINO against other models on long-tailed findings grouped by frequency: Common ($freq>10\%$), Uncommon ($1\%<freq<10\%$) and Rare ($freq<1\%$).
    \textbf{b)} Class distribution on the long-tail evaluations sorted by frequency.
    \textbf{c)} AUPRC comparisons on catheter malposition prediction.
    \textbf{d)} Macro Accuracy comparison on Normal vs Pneumothorax vs COVID-19 classification and ROC curves for the COVID-19 disease.
    \textbf{e)} Pearson's correlation coefficient comparison on Cobb angle regression using spinal exams on patients with scoliosis and the regression model's multi-head attention maps for explainability.
  }
  \label{fig:fig3}
\end{figure}

First, for radiology interpretation, we tested our method on labels with different frequencies and focus on the rare classes called the tail (MIMIC Long-Tail dataset~\cite{holste2023cxr}). We show results in Fig.~\ref{fig:fig3}a. This dataset with 26 classes contains very rare labels with less than 1\% of positive images. Details about classes' frequency are available in Fig.~\ref{fig:fig3}b.
We split them based on their prevalence in the training data and group them into three bins: Common ($freq>10\%$, $n=9$ classes), Uncommon ($1\%<freq<10\%$, $n=11$ classes), and Rare ($freq<1\%$, $n=6$ classes). We report macro AUPRC on each bin.
RayDINO improves over the other state-of-the-art methods for common classes ($57.0$ AUPRC, 95\%CI [$57.0$-$57.1$], $+0.9$ over UNIChest, $p\leq1\mathrm{e}{-5}$), but more interestingly our model significantly outperforms all competitors on uncommon diseases ($23.5$ AUPRC, 95\%CI [$23.5$-$23.6$], +$2.9$ over UNIChest, $p\leq1\mathrm{e}{-5}$) and is way ahead on rare classes ($22.4$ AUPRC, 95\%CI [$22.3$-$22.6$], +$6.2$ over ARK6, $p\leq1\mathrm{e}{-5}$). This showcases RayDINO's ability to capture rare clinical characteristics.

Then, we tested our method on catheter position prediction~\cite{Seah2020RANZCRCLiP}. We split the catheter categories (endotracheal tube, nasogastric tube and central venous catheter) into three groups based on the correctness of their position according to medical doctors: Normal ($freq=36.9\%$), Borderline ($freq=11.2\%$) and Abnormal ($freq=3.9\%$). Catheter malpositions happen rarely in clinical settings but are a severe source of complications for patients~\cite{pmid15217627}. This is reflected by the distribution of our dataset, where the Abnormal class is rare but is also the most important one. We present the results in Fig~\ref{fig:fig3}c. RayDINO outperforms alternatives by a significant margin on Normal ($94.3$ AUPRC, 95\%CI [$94.2$-$94.4$], $+5.2$ over ARK6, $p\leq1\mathrm{e}{-5}$), Borderline ($49.7$ AUPRC, 95\%CI [$49.4$-$50.0$], $+19.1$ over ARK6, $p\leq1\mathrm{e}{-5}$) and Abnormal catheters ($46.4$ AUPRC, 95\%CI [$44.9$-$47.9$], $+29.4$ over ARK6, $p\leq1\mathrm{e}{-5}$).
Notably, RayDINO achieves nearly triple the performances of the best alternative on abnormal positions, highlighting its effectiveness in identifying critical clinical scenarios.

\paragraph{Out-of-Domain Task Performance: Exploring New Diseases and Exams.}
All evaluations described above are related to diseases and exams that potentially occur in the training data of the foundation model, but more is needed in order to showcase a foundation model's capabilities.
First, in order to evaluate our model on a real out-of-distribution problem, we included a recent disease never seen by the model (using the POLCOVID dataset~\cite{suwalska2023polcovid}). We classify images between categories Normal, Pneumonia and COVID-19 and show results in Fig~\ref{fig:fig3}d and fig.~\ref{fig:cm_POLCOVID}. We observe that RayDINO reaches $78.0$ macro accuracy (95\%CI [$78.0$-$78.0$]). Our model significantly outperforms all other models ($+3.2$ over ARK6, $p\leq1\mathrm{e}{-5}$) and is even better for the COVID-19 disease reaching 91.0 AUROC (95\%CI [$90.0$-$92.0$], $+7$ over ARK6, $p\leq1\mathrm{e}{-5}$). 

We considered another out-of-distribution task by regressing spine curvature of 609 patients with scoliosis~\cite{wu2017automatic,wu2018unsupervised}. Radiography for spine curvature estimation differ from classical chest x-rays in term of field-of-view and aspect ratio and we used spinal X-rays that are a new exam not present in the training datasets.
We observe in Fig.~\ref{fig:fig3}e that RayDINO achieves a macro Pearson's correlation coefficient of $82.4$ (95\%CI [$82.1$-$82.7$]) across the 3 Cobb angles, clearly outperforming alternatives with $+15.1$ points over the best competitor ARK6 ($p\leq1\mathrm{e}{-5}$). We show in supplementary fig.~\ref{fig:scoliosis} that RayDINO's predictions are better for the main thoracic (MT) angle with a $r^2$ of $0.90$ while being less precise for the proximal thoracic (PT) and thoraco-lumbar (TL/L) angles. RayDINO achieves strong performances despite the fact that the field-of-view differs greatly from classical chest x-rays and that the regression adapter is trained on few hundreds of patient.

\begin{figure}[p]
    \centering
    \includegraphics{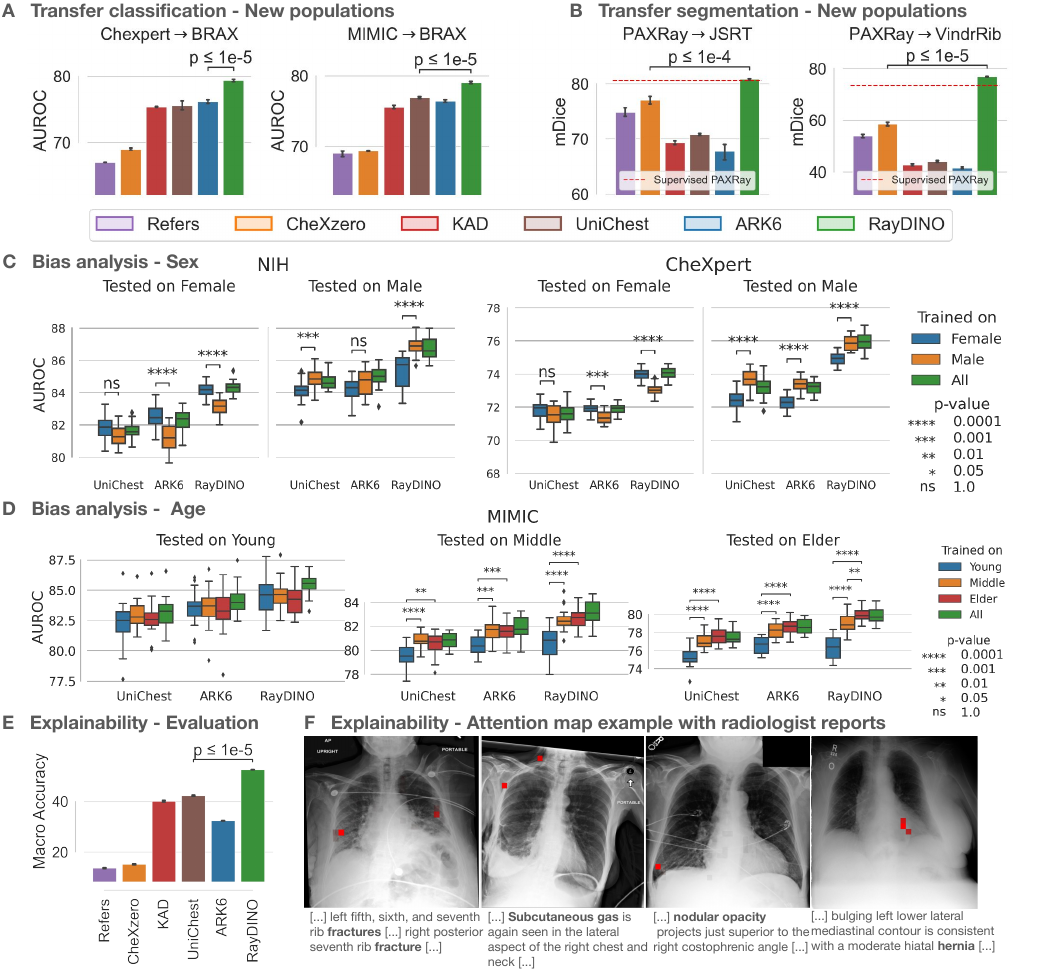}
    \caption{
        \textbf{Interpretability and Fairness evaluation of RayDINO between different demographics.}
        \textbf{a)} AUROC comparisons by training and validating a classifier on MIMIC or CheXpert (USA) and testing it on BRAX (Brazil). 
        \textbf{b)} mDice comparisons by training and validating a segmentation head on PAXRay++, a synthetic dataset generated from 2D CT-scan projections (mainly the USA and China), and testing it on real world datasets JSRT (Japan) and Vindr Rib (Vietnam). 
        \textbf{c)} AUROC fairness comparisons by training a classifier on one sex-specific split and testing it on the other split of NIH and CheXpert datasets. p-values obtained with a Mann-Whitney test to assess performance disparities when training with the same sex versus a different sex than the test set.
        \textbf{d)} AUROC fairness comparisons by training a classifier on one age-specific split and testing it on the two other splits of the MIMIC datasets.
        \textbf{e)} Macro Accuracy explainability comparisons by training one classifier for each class of Vindr and by looking if the position where the attention is looking the most match the radiologist bounding box on the test set. 
        \textbf{f)} Attention maps comparison with radiologist reports to demonstrate the accurate localization of RayDINO's explainable classifiers (specific finding highlighted in bold in the report).
    }
    \label{fig:fig4}
\end{figure}

\paragraph{Robustness to Patient Demographics and Superior Generalization in Clinical Settings.}
In clinical practice, it is crucial for models to be robust to the demographic and social origins of patients. Thus, we evaluated our classification models on external patients from populations with different profiles. We used either CheXpert or MIMIC as internal training data, both collected in the USA, and tested them on the external BRAX dataset~\cite{reis2022brax} from Brazil.
As depicted in Fig.~\ref{fig:fig4}a, RayDINO significantly outperforms all other models on the BRAX dataset when trained on CheXpert ($79.2$ AUROC 95\%CI [$78.8$-$79.6$], $+3$ over ARK6, $p\leq1\mathrm{e}{-5}$) and on Mimic ($79.0$ AUROC 95\%CI [$78.8$-$79.3$], $+2.1$ over UniChest, $p\leq1\mathrm{e}{-5}$). This demonstrates its robustness to different disease distributions and patient origins. We highlight an interesting observation in supplementary fig.~\ref{fig:scatter_transfer}: despite not being the top performer on the CheXpert dataset, RayDINO generalizes significantly better than all other models on a new population, indicating its ability to avoid bias and overfitting to the training population.

We further evaluate RayDINO's generalization performance in segmentation tasks, training on CT-scan projections mainly from the USA and China (PAXRay++) and testing on two external datasets from Japan (JSRT) and Vietnam (Vindr Rib), as shown in Fig.~\ref{fig:fig4}b. We emphasize the fact that the training and validation data are synthetic 2D chest X-rays made from CT-scan projections while the test data are real chest X-rays. Despite the synthetic nature of the training data and the distinct geographical origins of the patients, RayDINO shows significant generalization improvement with respectively $80.8$ and $76.9$ mDice on Japanese and Vietnamese patients (95\%CI [$80.5$-$81.0$] and [$76.7$-$77.1$], $+3.8$ and $+18.4$ over CheXzero, $p\leq1\mathrm{e}{-4}$ and $p\leq1\mathrm{e}{-5}$). RayDINO even outperforms the fully supervised CXAS model~\cite{Seibold_2023_CXAS} trained on PAXRay++ when evaluated on Vinr Rib with $+3.5$ mDice. We show an example of how both models segment real world images in supplementary fig.~\ref{fig:generalization_segmentation}.

\paragraph{Examining Performance Across Sex and Age Groups.}
Every foundational model must undergo a rigorous analysis of its learned biases to ensure fairness and reliability across diverse applications and populations. While we make no definitive claims about our model, this analysis is included to aid researchers in effectively utilizing our method.

First we evaluated sex-specific bias by following Larrazabal et al.'s setting~\cite{doi:10.1073/pnas.1919012117}. We assessed the AUROC difference on a test split comprising only one sex when training on the same sex versus training on the other sex (male or female, intersex and hermaphrodite data being unavailable). We use 20 folds and also included a combined sex split termed `All' as a topline. We report results in Fig.~\ref{fig:fig4}c.
According to the Mann-Whitney test, we observe that RayDINO exhibits a significant performance gap between male and female when tested on the same sex ($p\leq1\mathrm{e}{-4}$ for all four comparisons). The two other models, ARK6 and UniChest, show a non-significant gap in similar scenarios one time out of four for ARK6 and two times for UniChest, suggesting that self-supervised learning may be more sensitive to sex-specific characteristics in the training data.

However, while the gap might be more significant for RayDINO, it outperforms by a large margin other models in terms of raw performance on both NIH and CheXpert (respectively $+1.0$ and $+2.5$ AUROC on females $\rightarrow$ males with $p\leq0.01$ and $p\leq1\mathrm{e}{-5}$, $+1.8$ and $+1.6$ AUROC on males $\rightarrow$ females with $p\leq1\mathrm{e}{-5}$ and $p\leq1\mathrm{e}{-5}$). Notably, RayDINO surpasses these models when comparing its worst minority group performance against the best group performances of ARK6 and UniChest ($+0.6$ when testing on males and training RayDINO on females against the other best model trained on males, $p\leq0.05$, and $+1.1$ when testing on females and training RayDINO on males against the other best model trained on females with $p\leq1\mathrm{e}{-5}$). This indicates that even when the task-specific adapter is trained on biased data, RayDINO can effectively mitigate this bias. Despite a larger performance gap between males and females, RayDINO makes fewer errors on all splits.

To emulate a complex scenario where populations vary in terms of prevalence and visual characteristics, we extended this analysis to patient's age. We divided the MIMIC dataset into three age groups: Young/Y (0-35 years), Middle/M (35-60 years), and Elder/E (60-100 years), and sampled them equally. As depicted in Fig.~\ref{fig:fig4}d, RayDINO achieves the best performances across all combinations with an average AUROC increase of $+0.8$. It either matches the performance (in the Y$\rightarrow$M, Y$\rightarrow$E, E$\rightarrow$Y combinations with $p=0.47$, $p=0.62$ and $p=0.08$) or consistently outperforms the best alternative model ARK6 in the six other combinations ($p\leq0.05$). Further details about split and dataset statistics used for the bias analysis are provided in the Additional Results section and supplementary table~\ref{tab:data_stat}.

\paragraph{Explainability.}
Our model is designed to be inherently interpretable, utilizing a simple two-layer decoder with attention pooling on top of our localized holistic features. This enables the generation of attention maps, visual representations that highlight the specific areas the model focuses on during prediction. Examples of such attention maps, confirmed by expert radiologists from this study, are shown in Fig.~\ref{fig:fig2}c for Pneumothorax classification, in Fig.~\ref{fig:fig3}e for Scoliosis angle regression and in Fig.~\ref{fig:fig4}f for Fractures, Subcutaneous Emphysema, Nodules and Hernia. We show additional images in fig.~\ref{fig:explainability_supp} as well as an example of a identified failure cases thanks to RayDINO's explainability in fig.~\ref{fig:explainability_supp}.
By analyzing these maps, we can accurately pinpoint the location and extent of diseases within the images. Radiologists can also gain insights into the model's prediction process. 
For instance, in the scoliosis case, each attention head (represented by different colors) focuses on a different part of the image.
Such detailed decomposition of the model's reasoning process is not possible with traditional methods like GradCAM~\cite{DBLP:journals/corr/SelvarajuDVCPB16}, which only produce a single, less precise heatmap~\cite{doi:10.1148/ryai.2021200267}. 
We evaluated the pinpointing performance of each model by examining if the area with the highest attention score for each disease was within the radiologists' bounding boxes on the VinDr dataset (Fig.~\ref{fig:fig4}e). RayDINO outperforms all other model in term of localization, achieving a macro accuracy of $52.6$\% (95\%CI [$52.5$-$52.7$], $+10.3$\% over UniChest, $p\leq1\mathrm{e}{-5}$).
This level of interpretability not only offers insights into the model's decision-making process but also aids in validating its predictions, thereby enhancing trust in the model's outputs. 

\section*{Opportunities and challenges of RayDINO}
We developed RayDINO, the largest and strongest radiology model that can analyze chest X-rays in a holistic manner, and is trained in a patient-centric way using only the imaging data rather than annotations that follow country-specific guidelines. 
By using the most recent techniques for scaling self-supervision based on DINOv2, we present a model that has the potential to affect real clinical practice. Since X-rays is one of the most common medical examinations worldwide for screening and diagnosis particularly for chest applications, our model could be applied in places where access to healthcare is difficult and has potential impact on the lives of millions of people. In this section, we detail all the advantages of the RayDINO model for a comprehensive radiology interpretation. Moreover, we discuss challenges and limitations with respect to its wide use.

\paragraph{Impressive performances with respect to other foundation methods and the added value of RayDINO.} Recently, there are several models presented in the literature that propose general representations of X-ray imaging using either imaging, or coupling imaging with text, evaluated on different tasks~\cite{zhou2022generalized,tiu2022expert,dai2023unichest,zhang2023knowledge,ma2023foundation}. RayDINO has proven superiority with respect to these models. In addition, it has been assessed on more holistic and clinically relevant issues compared to other methods for X-ray image processing (REMEDIS~\cite{Azizi2023}, Rad-Dino~\cite{perez2024rad}), showing remarkable adaptation to unseen tasks, with few labels and good localization, staying robust on new populations and mitigating bias inherent to the data. To the best of our knowledge, our study is the first to train the largest  self-supervised method on 873k X-rays and test it across 9 tasks on 21 different benchmarks comprising more than 200 categories in both internal and external settings, from patients of seven countries across four continents, and reaching good explainability with state-of-the-art results in the process. Most articles focus on populations from one country or without proper external validation~\cite{Azizi2023,Zhou2023,perez2024rad}, but our extensive evaluation allows us to audit RayDINO's ability to handle multiple tasks across diverse demographics.

\paragraph{Holistic understanding based only on imaging and clinical applicability.} This study highlights multiple advances of image-based self-supervised foundation models, providing evidence for their holistic understanding and real clinical use. One of the first advances of RayDINO compared to other task-oriented models is that it does not make assumptions about specific classes. Instead, RayDINO focus on all discriminative characteristics of the X-ray, capturing the real properties of the patient without guideline-related limitations. This holistic approach allows RayDINO to tackle multiple tasks at once and to reach superior performances compared to other models on new diseases that do not exist in the original data such as COVID-19 or abnormalities such as pneumoperitoneum, pneumomediastinum and subcutaneous emphysema, from the MIMIC Long-Tail evaluation, which are clinically rare~\cite{pmid18980688,pmid26730086}.

Moreover, the general purpose for clinical use of our model is also highlighted in the report generation task. A radiologist's report includes multiple pieces of information such as whether a disease is present along with elements supporting that conclusion, as well as other general observations unrelated to the aforementioned disease, e.g. bone-related issues, presence of catheters or pacemakers, and so on. The variety of information included in text reports is difficult to cover with weakly-annotated datasets, justifying the new usage of textual supervision in models such as Refers~\cite{zhou2022generalized}, CheXzero~\cite{tiu2022expert} and UniChest~\cite{dai2023unichest}. Interestingly, RayDINO outperforms all of these models on the report generation task without using text annotations during its pretraining. These findings highlight that image-based models that are trained to explicitly search for discriminative or rare features, thanks to self-supervised learning, can outperform multimodal models when building open-ended medical artificial intelligence~\cite{moor2023foundation}.

\paragraph{Applicability and generalization across geographical, demographic and biological characteristics.}
While defining fairness in medical imaging remains a complex challenge~\cite{RicciLara2022, Bernhardt2022}, RayDINO has the potential to reduce disparities in access to medical care across diverse populations. Studies have highlighted variations in medical practice within the United States, with clinicians demonstrating differences in the treatment and diagnosis of disorders among patients from different demographics, irrespective of the patients' actual needs~\cite{doi:10.1148/radiol.2021203742,10.1001/jama.2021.13304,doi:10.2214/AJR.17.18907}. Moreover, diagnostic guidelines can vary between countries, but our model aims to cater to a broader audience beyond the countries it has been trained on. 
By not relying on human annotations for training, our RayDINO backbone has learned none of the bias inherent in patient-based diagnostic guidelines and radiologists’ interpretations.
This also removes errors that can arise in automatic annotation pipelines~\cite{irvin2019chexpert,johnson2019mimic,Bustos_2020}, enabling the fairer training of RayDINO and the extraction of holistic features that accurately represent the patient's condition.

However, such bias can re-emerge during the fine-tuning of the foundation model using annotations. To address this, we evaluated our backbone in a frozen setting, without modifying its parameters, and instead trained very small task-specific adapters. This approach offers three key advantages. Firstly, using the same foundation model for all tasks enhances the ability to audit the raw features of the model without incorporating label and task-specific bias. Secondly, the bias from the annotations is captured only by a simple decoder, reducing overfitting, as demonstrated in our generalization evaluation from the USA to Brazil. Lastly, the inherent interpretability of our adapters allows radiologists to directly verify whether the model is relying on spurious correlations in the X-ray for its predictions. The consistent and significant improvement of RayDINO on the worst minority groups with respect to sex and age, as well as the significant classification transfer on new populations, shows the potential of this model for clinical use and adaptability in regions with limited access to healthcare systems.

\paragraph{Limitations.}
Although RayDINO performs significantly better than prior work, it isn't devoid of limitations. Firstly, to enhance generalization to external populations, future iterations should incorporate data from a broader range of regions and demographics beyond the USA and Europe and test on external populations other than Brazil, Vietnam and Japan. Secondly, while RayDINO demonstrates improved performance for minority groups and bias mitigation, it exhibits a slightly larger gap between performances stratified by sex. 
Thirdly, the group-specific bias could be developed to provide more comprehensive results. This could be achieved by exerting more control over the studied population, for instance, through an intersectionality analysis. Finally, the primary objective of this study is to explore the capabilities of our unimodal foundation model, which is solely based on 2D imaging. We aimed to demonstrate its ability to perform a holistic analysis of X-rays, with a particular emphasis on diagnostic tasks. Future work will focus on using this model on prognostic tasks investigating also the importance of patient's history, symptoms and follow-up as well as 3D CT-scans.

\bibliography{egbib}

\bibliographystyle{Science}

\section*{Acknowledgments}
We would like to thank Mathilde Caron for initial discussion around this project. 
We would also like to thank Timothée Darcet, Camille Couprie, Huy V. Vo, Federico Baldassare, Marc Szafraniec, Vasil Khalidov, Michaël Ramamonjisoa, Cijo Jose, Seungeun Yi, Patrick Labatut for their help and support and Naila Murray for her valuable comments.

\paragraph{Funding} This work was partially supported by the ANR Hagnodice ANR-21-CE45-0007.

\paragraph{Authors contributions}
T.M., A.J., P.B., M.O. and M.V. conceived and designed the study. T.M. developed the deep learning model and the evaluation code, performed the statistical analyses and made the visualization. M.M. and M.O. provided intellectual advice to develop the deep learning model. Y.L. and C.H. provided guidance and supported the research. G.C. and M.-P.R. contributed to the clinical interpretation of X-rays and validated the clinical usefulness of the evaluation tasks. P.B., M.O. and M.V. supervised the study. T.M., P.B., M.O. and M.V. wrote the manuscript, and many authors have provided valuable edits.

\paragraph{Competing interests}
No competing interests.

\paragraph{Data and materials availability}
All data are available in the manuscript or the supplementary materials. Training code available at \cite{gitdinov2}. Models will be available soon.

\section*{Supplementary Materials}
Materials and Methods \\
Additional Results \\
Figs. S1 to S7 \\
Tables S1 to S2 \\
References (49–56) \\

\newpage

\setcounter{figure}{0}
\setcounter{table}{0}
\renewcommand{\figurename}{fig.}
\renewcommand{\thefigure}{S\arabic{figure}}
\renewcommand{\tablename}{table}
\renewcommand{\thetable}{S\arabic{table}}

\section*{Materials and Methods}

\subsection*{Unsupervised Pretraining}
Throughout this work, we use Vision Transformers~\cite{dosovitskiy2020image} (ViTs).
These models, when given a new image, produce a global representation, as well as a feature map describing individual patches of 16x16 pixels in the context of the image.
For a 224x224-pixel image, the model provides one global representation token \verb+[CLS]+, and a 14x14 map of patch-level representation tokens.
We train our RayDINO models using an unsupervised feature learning method, adapting the training algorithm of DINOv2~\cite{oquab2024dinov}. 
The training procedure proposed in that work uses three different loss functions:
(i) the DINO loss, first proposed in DINO~\cite{caron2021emerging}, encourages different data augmentation views (e.g. crops, geometric and colorimetric distortions) of a given image to lead to the same \verb+[CLS]+ token.
(ii) a regularization term, KoLeo~\cite{sablayrolles2019spreading}, ensuring better coverage of the representation space by pushing the \verb+[CLS]+ tokens of different images away from each other.
(iii) the iBOT loss, first proposed in iBOT~\cite{zhou2021ibot}, allows learning better local representations through masked image modeling: during training, the model learns to predict a hidden part of its input, based on the surrounding context.

Following previous work, we train the biggest model currently available for radiology based on the ViT architecture with 307M parameters coined ViT-L (Large), producing tokens of dimensionality 1024. 
We provide an ablation of the performance as a function of model size in Fig.~\ref{fig:size}, showing in particular that performance increases with model size.
We train the model for 125k iterations with a minibatch size of 2048. 
The learning rate follows a standard cosine schedule with 12.5k iterations of linear warmup to a peak value of $1.10^{-3}$. 
The weight decay follows a cosine schedule from $0.04$ to $0.4$. 
We point out that these are the default hyperparameters of the "Fast Setup" setting in the open-source DINOv2 implementation\footnote{\url{https://github.com/facebookresearch/dinov2}}. 
We deviate from the default settings only for the image crops due to the higher level of detail of XRay images: we employ 512x512-pixel crops (instead of 224x224) for the large crops, 224x224 (instead of 96x96) for the small crops; additionally, the large crops must cover \textit{at least} 15\% of the image (instead of 32\%) while the small crops cover \textit{at most} 15\% of the image (instead of 32\%). 
Training is performed on 128 A100-80GB Nvidia GPUs and lasts 36 hours.

\subsection*{Data}
The original DINOv2 model was trained on a dataset composed of 142M images, curated from a large pool of internet images.
However, in the context of XRay image analysis, we don't have access to such large web-sized data pools: therefore we combine four publicly available datasets (CheXpert, MIMIC, NIH-14, PadChest) to train the RayDINO models. 
This sums up to approximately 863k images used for our unsupervised pretraining; in contrast to supervised approaches, we do not employ any annotations; therefore, it is possible to increase the size of the pretraining dataset easily by simply collecting more radiographs or including more suitable datasets. 
We use train, validation and test splits for all evaluations. To make our results reproducible we used all official test splits when available. When no such official splits are available we used respectively 80\%, 10\% and 10\% of the data for the train, validation and test sets. We used the train set as a validation set for JSRT as the number of images is small.

The evaluation procedure for segmentation is run on the following datasets and segmentation labels:
JSRT~\cite{shiraishi2000development} is annotated with lungs, heart, clavicles. Vindr Rib~\cite{nguyen2021vindrribcxr} evaluates on the 20 individual ribs. PAXRay++~\cite{Seibold_2022_BMVC, Seibold_2023_CXAS} contains 150 classes including soft organs such as heart, liver, lungs, as well as bones such as individual ribs and vertebrae. Finally, Pneumothorax~\cite{zawacki2019siimacr} proposes a pneumothorax disease segmentation task.

\subsection*{Comparisons \& Baselines}
In all our evaluations, we compare our model to several publicly available X-ray foundation models. 
We select five recent methods, representative of the current state of the art, as comparison baselines.
REFERS~\cite{zhou2022generalized} is a model trained using cross-supervision between radiology images and text from radiology reports.
CheXzero~\cite{tiu2022expert} employs contrastive language-image pretraining (akin to CLIP~\cite{radford2021learning}) to produce a vision model able to tackle Chest X-ray images.
KAD~\cite{zhang2023knowledge} is similarly trained using paired radiology images and reports, with the addition of a knowledge-base component to improve performance. Inference is then performed following the CLIP zero-shot protocol, effectively performing classification over the list of extracted disease names. UNIChest~\cite{dai2023unichest} consists of an improved model built on top of the KAD method.
Finally, ARK6~\cite{ma2023foundation} is a foundation model trained on a concatenation of multiple existing datasets used in a classification training pipeline, effectively using a large amount of categorical supervision.
Regarding training data, the models against which we show comparisons are trained on a relatively similar set of datasets as our model.
REFERS, CheXzero, and KAD are trained on the Mimic dataset~\cite{johnson2019mimic}.
UniChest additionally leverages NIH~\cite{wang17nih}, CheXpert~\cite{irvin2019chexpert}, and Vindr~\cite{nguyen2020vindrcxr}.
ARK6 uses all these datasets along with the data from the RSNA Pneumonia Detection Challenge~\footnote{\url{https://www.kaggle.com/c/rsna-pneumonia-detection-challenge/}} and the Shenzen dataset~\cite{jaeger2014two}.

Recent concurrent work RAD-DINO~\cite{perez2024rad} explores a direction similar to our approach by building on DINOv2 models. 
They initialize a vision model with a DINOv2 checkpoint, using a smaller model ViT-B of 86M parameters, then perform training for a small number of iterations (7 times less than RayDINO) using a combination of datasets and compare favorably against smaller models found in the literature.
They find that text supervision is not necessary to build strong vision models for the medical domain, that SSL features present viable properties for medical applications, and that it should be possible to build a strong foundational image encoder with an SSL approach.
In this study, our model RayDINO successfully tests that hypothesis, leading to a versatile model that outperforms all prior state-of-the-art approaches on a multitude of downstream tasks. We compare our work against the strongest model available in the literature on a wider variety of tasks including very important ones like external validation on new demographics and age and sex-related bias analysis. On top of that, RAD-DINO is closed source while our model is openly available.

\subsection*{Experimental Protocols}

\subsubsection*{Classification}
All the tasks considered in this part for the evaluation take as input the whole image and have a simple output space (classification or regression).
As mentioned before, for all models, we keep the parameters of the image encoder frozen and learn a simple decoder on top using stochastic gradient descent.
Thanks to the use of very small decoders, we can use the validation set to select the best decoder through a grid search over the following hyperparameters:
learning rate: 13 values: {\small$[1.10^{-5}$, $2.10^{-5}$, $5.10^{-5}$, $1.10^{-4}$, $2.10^{-4}$, $5.10^{-4}$, $1.10^{-3}$, $2.10^{-3}$, $5.10^{-3}$, $1.10^{-2}$, $2.10^{-2}$, $5.10^{-2}$, $1.10^{-1}]$}; weight decay: 3 values: {\small[$0$,  $1.10^{-5}$,  $1.10^{-4}$]}; number of layers in decoder: 1 (linear) or 2 (MLP); feature pooling: attentive pooling using $n_{classes}$ heads or average pooling. In total, the grid contains 156 hyperparameter combinations that we execute at a minimal cost, by performing only a single inference of the image encoder and training decoders simultaneously.

We use a binary cross-entropy loss for multi-label classification tasks (CheXpert~\cite{irvin2019chexpert}, MIMIC~\cite{johnson2019mimic}, NIH~\cite{wang17nih}), a simple cross-entropy loss for POLCOVID (as it is not multi-label), and the SMAPE loss for Scoliosis regression.
All experiments are replicated on 5 random seeds to produce error bars.

\subsubsection*{Segmentation}
We evaluate all models on 4 datasets: PAXRay++~\cite{Seibold_2022_BMVC, Seibold_2023_CXAS}, JSRT~\cite{shiraishi2000development}, Vindr Rib~\cite{nguyen2021vindrribcxr} and SIIM Pneumothorax~\cite{zawacki2019siimacr}. We compare the models using the macro averaged Dice score computed per class. We use the official sets from Vindr Rib, take 80/10/10\% splits for PAXRay and 80/20\% splits for JSRT by validating on the train set, and use the stage 2 validation data from Siim Pneumothorax as test set. All images are rescaled to $512$x$512$ pixels for RayDINO and the nearest smaller multiple of the \texttt{patch\_size} for other models and metrics are computed at $512$x$512$ pixels.
All backbones are frozen and we train conjointly a UperNet head and a FCN head with a feature pyramid network on top of the extracted features using a Dice loss. We train for 5000 iterations using a batch size of 16 and run a grid search over 9 learning rates [{\small$1.10^{-5}$, $5.10^{-4}$, $1.10^{-4}$, $1.10^{-3}$, $3.10^{-3}$, $1.10^{-2}$, $3.10^{-2}$, $1.10^{-1}$, $3.10^{-1}$}] with a linear schedule (1500 warmup iterations, then linear decay until 0).

\subsubsection*{Generalization}

Models are evaluated on two generalization tasks. For the population generalization tasks we select the classification heads trained and validated on CheXpert~\cite{irvin2019chexpert} (USA, 224k images) or MIMIC~\cite{johnson2019mimic} (USA, 377k images), and evaluate them on the whole BRAX~\cite{reis2022brax} dataset (Brazil, 41k images), reporting AUROC. For the population and CT-scan projection generalization we train and validate segmentation heads on PAXRay++~\cite{Seibold_2022_BMVC,Seibold_2023_CXAS} (multiple countries, mainly China and the USA, 15k images) and test them on JSRT~\cite{shiraishi2000development} (Japan, 154 images) and Vindr Rib~\cite{nguyen2021vindrribcxr} (Vietnam, 245 images) by mapping labels accordingly and reporting mDice score.

\subsubsection*{Report generation}

We finetune a publicly available large language model, namely the LLama2-7B model~\cite{touvron2023llama} to perform report generation. We first convert image features, using a small two-layer Transformer adapter, into tokens of the dimensionality expected by the language model. Then, this language model can be trained to produce reports directly from the image tokens. To this end, we employ an image captioning loss to generate the findings section of the MIMIC~\cite{johnson2019mimic} reports, using the indication section as prompt when available. We train for 10.000 iterations with a batch size of 8 sequences of 2048 tokens.
For fair comparison, we align on the split and the text preprocessing used in MAIRA-1~\cite{hyland2023maira}, and refer to their manuscript for more details.

\subsubsection*{Bias analysis}
To perform the bias analysis of our RayDINO backbone, we train classification heads using only subsets of the same size from the training data corresponding to a category of interest, e.g. the \textit{female} subset in NIH~\cite{wang17nih}. Then, we evaluate the resulting model on the other categories to assess the presence of bias, and report the AUROC metric. We also train a classification head on the whole training data of all subsets to observe the performance differences with respect to represented categories. For each task and data subset in the bias analysis, we follow the protocol for classification evaluation described further above. We repeat this on 20 different folds and report the results of a Mann-Whitney test to look for significant differences.

\subsubsection*{Explainability}
We train a disease-specific classification adapter for each model using a single attention head on the VinDr training set. We then generate attention maps for each test image in the VinDr dataset. To calculate the final accuracy result for each disease, we examine all true disease bounding boxes and check if the patch with the highest attention score from the specific disease's adapter intersects with the bounding box by at least half the patch's area. We then compute the macro average of accuracies for all diseases and report the result.

\subsection*{Evaluation Metrics}
Throughout this work, we employ several evaluation metrics to compare our model to the state of the art in the literature, and we describe these here for the case of image-related tasks.
For multilabel binary classification, we employ the Area Under the Receiver-Operator Curve (AUROC) metric, and report the macro average over classes.
For long-tail binary classification, we employ the Area Under the Precision-Recall Curve (AUPRC), as this metric handles unbalanced classes better, and we report the macro averaged AUPRC over classes, when applicable.
For simple classification (where classes are mutually exclusive, as in POLCOVID~\cite{suwalska2023polcovid}), we report the accuracy of the classifier averaged over classes (Macro Accuracy).
For segmentation tasks, we use the Dice metric, and report the average over classes (mDice); this metric indicates the quality of overlap between predicted and ground truth segmentation masks.
For regression tasks (e.g. scoliosis angle), we report the correlation value between the prediction and the ground truth.

For evaluating text outputs, as done for the report generation task, the performance measurement is more complex. 
In order to compare with alternatives, we report several metrics used in the literature, namely BLEU-4~\cite{papineni2002bleu}, ROUGE-L~\cite{lin2004rouge}, RadGraph F1, CheXbert vector, RadCliQ, and Macro-F1-14.
BLEU-4 and ROUGE-L are metrics designed for the evaluation of translation models, and allow measuring the quality of generated text with respect to a reference text. In the case of report generation, they are approximate metrics that we employ as proxies for reference, using the \texttt{pycocoevalcap} library.
For proper comparison with the MAIRA-1 work, proposing a radiology report generation pipeline, we also compute the same metrics as their authors and refer to their manuscript for additional details.
RadGraph-F1 is a metric based on the RadGraph model, which parses generated and ground truth radiology reports into graphs containing clinical entities and relations between them, then computes the overlap which is averaged into a metric. RadCliQ combines RadGraph-F1 and BLEU. Similarly, the CheXbert vector metric compares a generated and reference report using the CheXbert model. CheXpert F1 metrics (referred to as Macro-F1-14 and Micro-F1-14 in the results) use the CheXbert automatic labeller to produce a metric based on the precision and recall over the 14 radiology observations. We used the reference implementation provided with the original work introducing the metrics\cite{yu2023evaluating}.
We multiply all metrics between $0$ and $1$ by $100$ to make them easier to read.

\subsection*{Statistical analysis}
When comparing the performance of different models, we employ the t-test to measure the statistical significance of the accuracy responses over a population of models trained with different random seeds, and assess the probability p that the difference in mean is due to chance. A low p-value means the result is statistically significant.
In the bias analysis, we train populations of models (using 20 different folds) using labels from different groups considered (e.g. age, sex) and obtain classification accuracy scores; we then employ the Mann-Whitney test to measure, by sampling X and Y from the two groups, that the probability of $X>Y$ is equal to the probability of $Y>X$. 
A high value suggests that populations perform equally.

\section*{Additional Results}
We provide additional experimental results and details that complement the analysis presented in the Results Section of the main text. 
We show additional metrics and comparison points for report generation,
statistics for the data used in the fairness analysis,
an analysis of the influence of model scaling and data source,
and additional analysis for the POLCOVID Classification, spine curvature estimation and Robustness to Patient Demographics probes. We also add more visualizations of RayDINO's explainability and a comparison of segmentation generalization from 2D CT-scan projection to real-world X-rays between RayDINO and a supervised segmentation model on PAXRay++ called CXAS~\cite{Seibold_2023_CXAS}.

\subsubsection*{Report generation}

We present in Table~\ref{tab:report} the evaluation results on report generation for all the competitor models considered, including the strong recent state-of-the-art MAIRA-1 model, specifically designed for this task; we provide 95\% confidence intervals on performance metrics for better comparison. We observe that RayDINO performs on par or signifiantly better than all competitor models on all metrics.
In particular, RayDINO significantly outperforms MAIRA-1 on radiologist-aligned metrics (RadGraph F1, CheXbert score, RadCliQ). We note that, interestingly, RayDINO performs on par with the strongly-engineered GPT4-augmented version MAIRA-1$^\dag$ on all metrics, while relying only the MIMIC-CXR reports.

\begin{table}[htb]
\centering
\resizebox{\textwidth}{!}{
\begin{tabular}{@{}l ccccccc@{}}
\toprule
Model & Bleu 4 & Rouge-L & RadGraph F1 & CheXbert score & RadCliQ $\downarrow$ & Macro-F1-14 & Micro-F1-14 \\
\midrule
\cg{$\text{MAIRA-1}^{\dag}$} & \cg{14.2}\ci{13.7}{14.7} & \cg{28.9}\ci{28.4}{29.4} & \cg{24.3}\ci{23.7}{24.8} & \cg{44.0}\ci{43.1}{44.9}  & \cg{3.10}\ci{3.07}{3.14} & \cg{38.6}\ci{37.1}{40.1} & \cg{55.7}\ci{54.7}{56.8} \\
\text{MAIRA-1 } & \textbf{14.0}\ci{13.6}{14.5} & 28.5\ci{28.0}{28.9} & 22.8\ci{22.1}{23.3} &42.8\ci{42.0}{43.6} & 3.16\ci{3.13}{3.19} & \textbf{36.7}\ci{35.0}{38.2} &  54.3\ci{53.3}{55.4} \\
\midrule
Refers & 10.8\ci{10.5}{11.3} & 27.3\ci{26.6}{27.9} & 19.8\ci{19.2}{20.5} & 36.7\ci{35.9}{37.5} & 3.36\ci{3.33}{3.40} & 23.2\ci{22.0}{24.6} & 40.1\ci{38.6}{41.3}  \\
CheXzero & 10.5\ci{10.2}{10.9} & 27.6\ci{27.0}{28.1} & 20.5\ci{19.9}{21.0} & 38.8\ci{37.9}{39.7} & 3.30\ci{3.27}{3.34} & 26.4\ci{24.9}{27.8} & 43.3\ci{42.1}{44.6}  \\
KAD & 12.7\ci{12.3}{13.1} & 29.6\ci{29.0}{30.2} & 23.0\ci{22.4}{23.7} & 42.8\ci{42.0}{43.7} & 3.13\ci{3.09}{3.17} & 32.5\ci{30.9}{34.1} & 50.2\ci{48.8}{51.4} \\
UniChest & 13.0\ci{12.5}{13.4} & 29.7\ci{29.1}{30.3} & 23.5\ci{23.0}{24.2} & 43.6\ci{42.8}{44.5} & 3.10\ci{3.07}{3.14} & 32.5\ci{30.9}{34.1} & 52.4\ci{51.1}{53.6}  \\
ARK6 & 12.5\ci{12.0}{12.9} & 29.3\ci{28.8}{30.0} & 22.8\ci{22.1}{23.4} & 42.8\ci{42.0}{43.7} & 3.14\ci{3.10}{3.18} & 32.2\ci{30.6}{33.5} & 51.1\ci{49.8}{52.1}  \\
RayDINO & 13.8\ci{13.4}{14.2} & \textbf{30.3}\ci{29.7}{30.8} & \textbf{23.9}\ci{23.3}{24.6} & \textbf{44.8}\ci{44.0}{45.6} & \textbf{3.07}\ci{3.04}{3.11} & 36.1\ci{34.4}{37.6} & \textbf{54.8}\ci{53.5}{56.0} \\
\bottomrule
\end{tabular}
}
\caption{
  \textbf{Report generation results:}. We report median and 95\% confidence
  intervals based on 500 bootstrap samples from the MIMIC-CXR test set. $\dag$ trained on GPT-4 augmented reports. RayDINO performs on par or significantly better than all competitors on all metrics.
}
\label{tab:report}
\end{table}

\subsubsection*{Scaling model and dataset size}
\begin{figure}[htb]
    \centering
    \includegraphics[width=\textwidth]{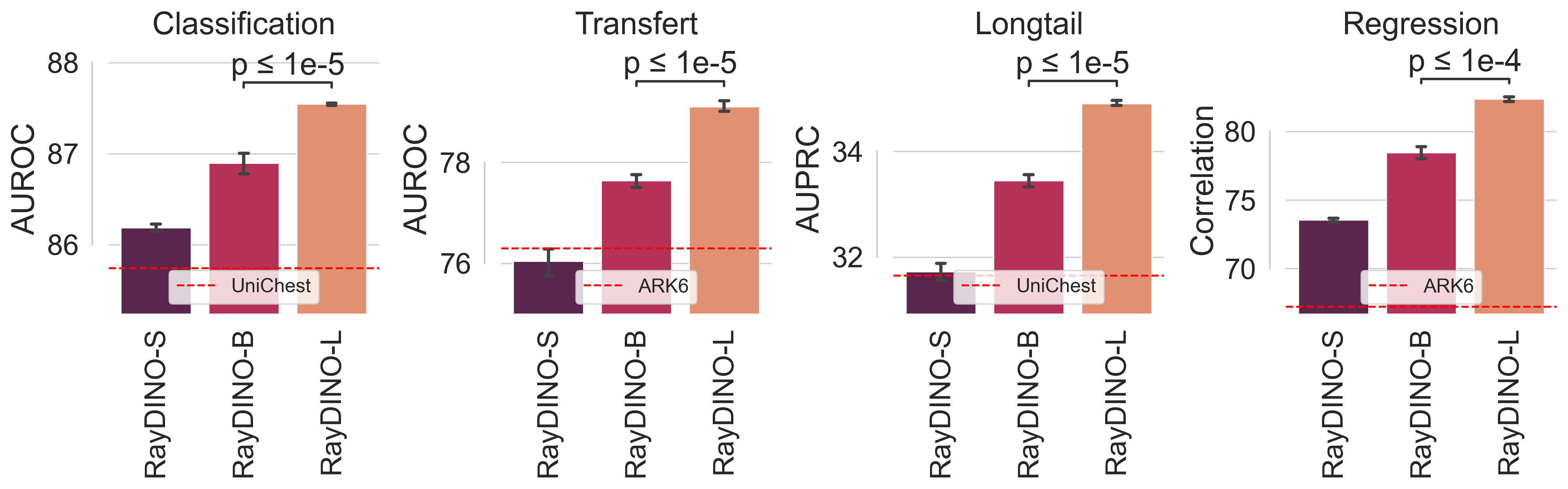} \\
    \vspace{1em}
    \includegraphics[width=\textwidth]{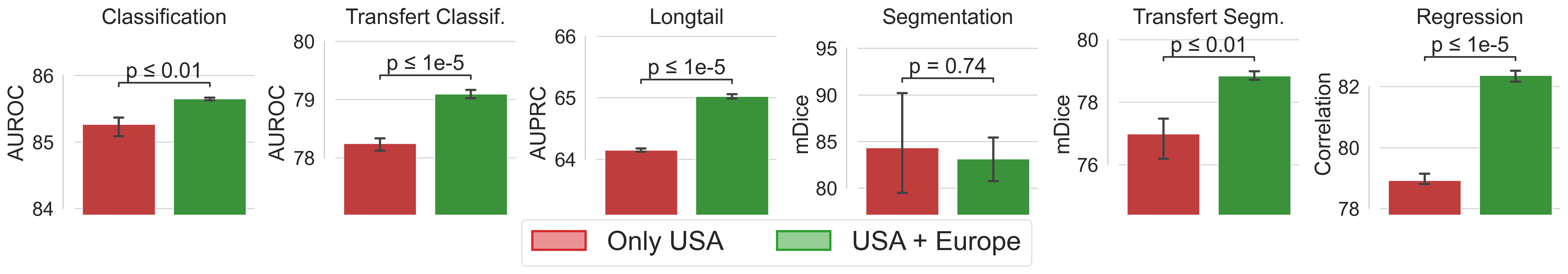}
    \caption{
        \textbf{(top)} Scaling Curve: we compare the impact of model's size on the aggregated results for multiple benchmarks. We report the best competitor model with a red dashed line as a reference.
        \textbf{(bottom)} Ablation on Populations from Training Data: we compare the impact of using patients from the USA only versus the USA and Europe when pretraining RayDINO.
    }
    \label{fig:size}
\end{figure}

We perform an ablation study of different model architecture sizes to assess the importance of model size in achieving strong performance. 
We test our approach on the standard ViT-S (21M params), ViT-B (86M params), ViT-L (307M params) architectures and present the results in fig.~\ref{fig:size}~\textbf{(top)}. 
For clarity and brevity, we aggregate tasks in groups. 
For each group of tasks we report the performance of best competitor model.
Unsurprisingly, we observe that the performance significantly increases with model size, indicating a healthy scaling trend.
More interestingly, the smaller models match or compare favorably with the best competitor model for all task groups, confirming the superior performance and versatility of the RayDINO pipeline regardless of the model size.

In a second ablation, we delve into the influence of the data source we use for training our models. 
RayDINO is trained on a combined dataset of images from both the USA and Europe, but to evaluate the impact of a diverse set of populations during the pretraining we train a similar model exclusively on images from the USA. Interestingly, we show in fig.~\ref{fig:size}~\textbf{(bottom)} that the models maintain consistent performance on traditional classification and segmentation tasks regardless of if the training dataset contains images only from the USA or if we also add European data ($-0.4$ AUROC, $p\leq0.05$ and $+1.2$ mDice, $p=0.74$ when we remove European data). This observation suggests that these conventional tasks may not fully capture the models' capabilities or their ability to generalize. However, the inclusion of a more diverse dataset in the training process significantly enhances the model's ability to generalize across different populations in classification ($+0.8$ AUROC, $p\leq1\mathrm{e}{-5}$), in segmentation ($+1.9$ mDice, $p\leq0.01$) and to more accurately predict rare classes ($+0.9$ AUPRC, $p\leq1\mathrm{e}{-5}$), thereby boosting its overall performance and applicability in real-world clinical scenarios. While it is systematically better or similar to include European data from PadChest, we point out that the absolute performance difference is of the order of one percent and the strong performances of our model is not only due to the additional data used.

\subsubsection*{Detailed Analysis for Out-of-Domain Task Performance}
\begin{figure}[htb]
    \centering
    \includegraphics[width=0.27\textwidth]{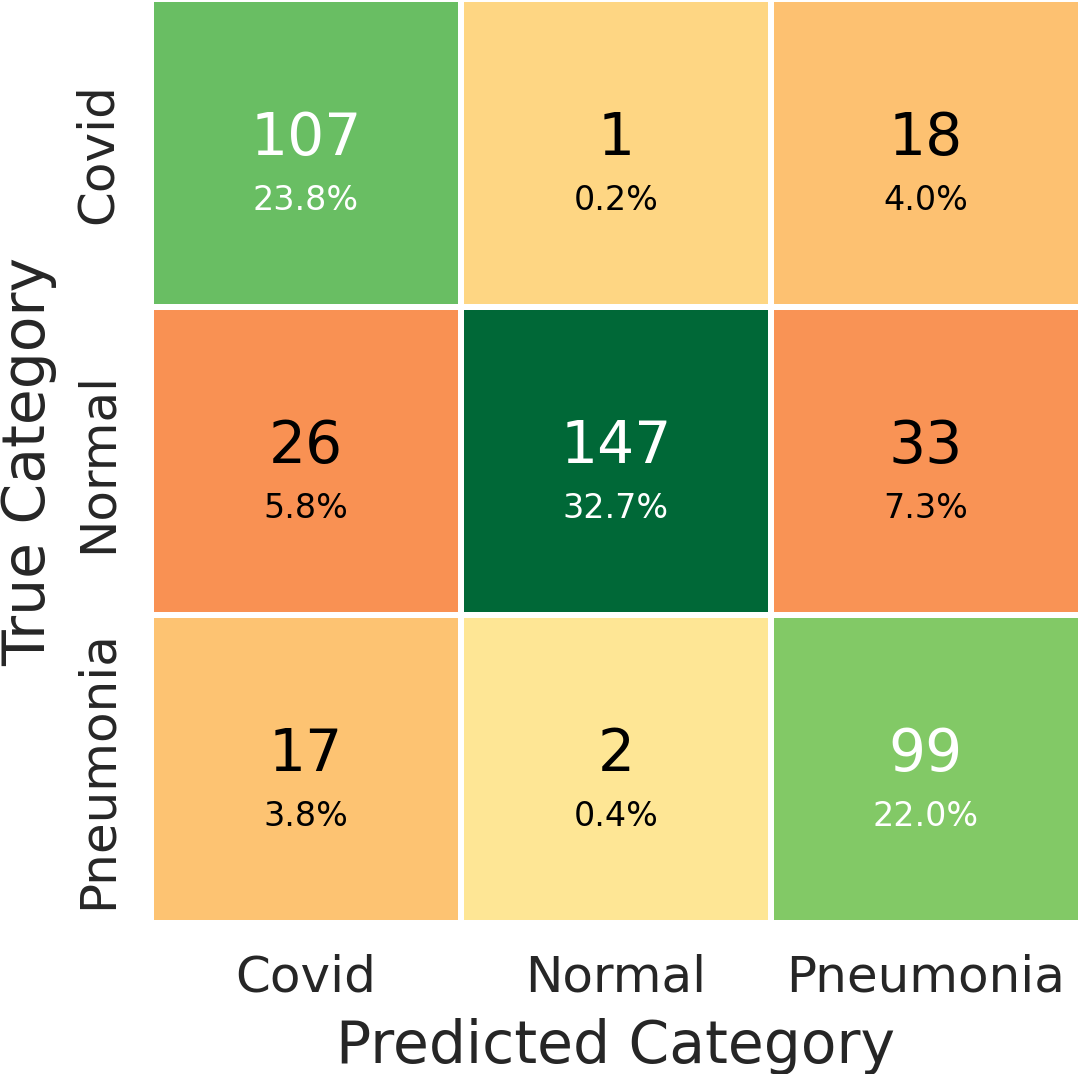}
    \caption{
        \textbf{POLCOVID Confusion Matrix:} we present the confusion matrix of RayDINO on the test set of the POLCOVID dataset.
    }
    \label{fig:cm_POLCOVID}
\end{figure}

\begin{figure}[htb]
    \centering
    \includegraphics[width=0.67\textwidth]{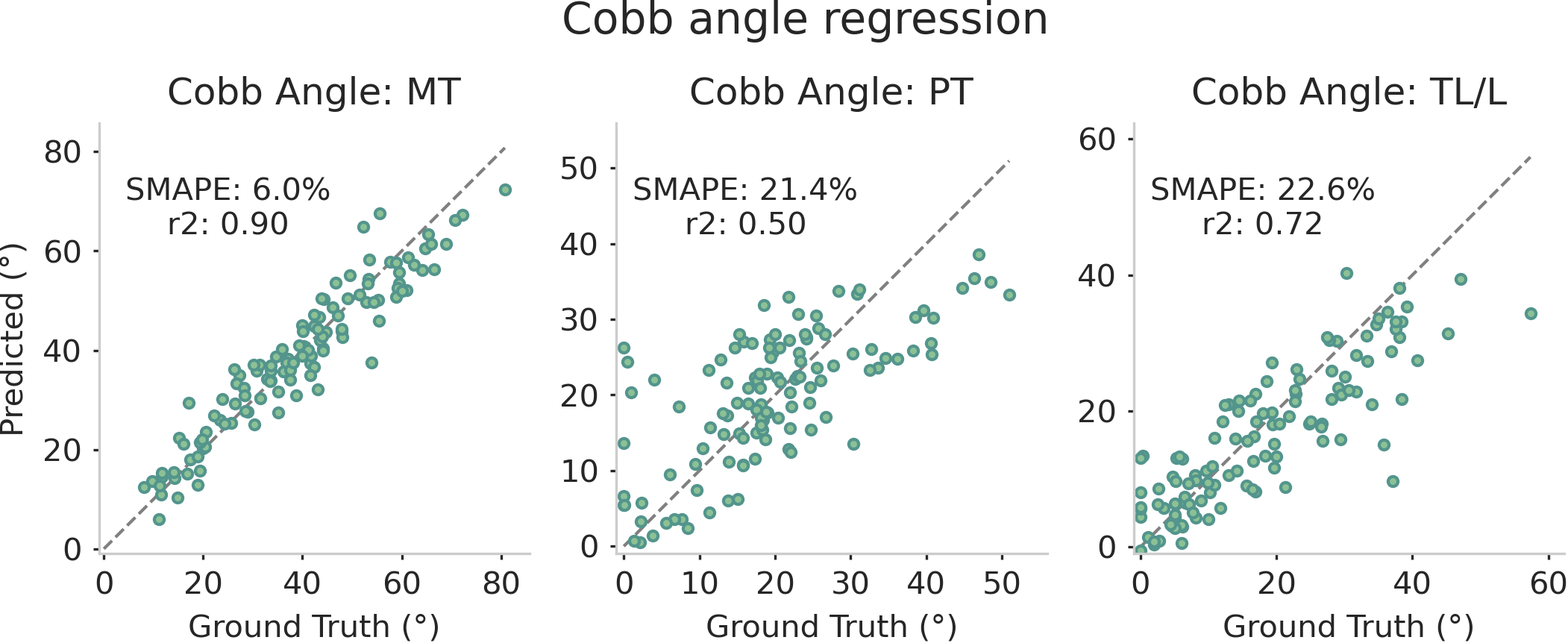}
    \caption{
        \textbf{Details of the scoliosis angle regression.} We report SMAPE and $r^2$ on the test set for main thoracic (MT), proximal thoracic (PT) and thoraco-lumbar (TL/L) angles.
    }
    \label{fig:scoliosis}
\end{figure}

We study more in detail the results obtained on out-of-domain tasks in fig.~\ref{fig:cm_POLCOVID} and fig.~\ref{fig:scoliosis}. On POLCOVID, the confusion matrix shows that the model predicts very few (0.6\%) false negatives on this dataset when considering the ill/healthy classification task. In a medical context, we want to avoid these type-II errors as much as possible, to make sure ill patients are treated. 
On the Scoliosis angle regression task, we observe that the model predictions are much more precise for the Main Thoracic curve Cobb angle (MT: $r^2=0.90$) than for the other angles (PT: $r^2=0.50$ and TL/L: $r^2=0.72$).

\subsubsection*{Statistics of the bias analysis dataset}

In table~\ref{tab:data_stat}, we present statistics on the dataset splits used for our bias analysis. We observe that there is a significant disparity in terms of patients with findings based on sex and age splits. For example, males have $9\%$ fewer findings than females in NIH and $11.8\%$ fewer findings in CheXpert, making these splits easier to predict. Similarly, younger patients have significantly fewer findings compared to the Middle and Elder splits ($-13.2\%$ and $-34.2\%$, respectively). This disparity may be due to the data sampling process and may not reflect actual clinical prevalence. It could also introduce bias in models trained on one split and tested on another, even if the model itself is not biased towards a particular sex or age group. Therefore, it is important to carefully consider the data sampling strategy during the generation of splits to ensure equal odds of finding-specific outcomes.

\begin{table}[htb]
    \centering
    \begin{tabular}{@{} c lccc @{}}
        \toprule
        & Split  & \female \ (\%) & Age & No Finding (\%) \\
        \midrule
        \parbox[t]{3mm}{\multirow{3}{*}{\rotatebox[origin=c]{90}{Mimic}}}
        & Young     & \phantom{0}57.4   & 27.1 \ {\small (\phantom{0}5.0)} & 68.8           \\
        & Middle    & \phantom{0}49.0   & 50.1 \ {\small (\phantom{0}6.8)} & 55.6           \\
        & Elder     & \phantom{0}52.9   & 74.6 \ {\small (\phantom{0}9.0)} & 34.6           \\
        \midrule
        \parbox[t]{3mm}{\multirow{2}{*}{\rotatebox[origin=c]{90}{NIH}}}
        & Female    & 100.0             & 46.8 \ {\small (16.3)} & 49.2               \\
        & Male      & \phantom{00}0.0   & 46.9 \ {\small (17.3)}  & 58.2               \\
        \midrule
        \parbox[t]{3mm}{\multirow{2}{*}{\rotatebox[origin=c]{90}{ChX.}}}
        & Female    & 100.0             & 62.4 \ {\small (19.0)}      & 22.5               \\
        & Male      & \phantom{00}0.0   & 59.5 \ {\small (18.4)}      & 34.3               \\
        \bottomrule
    \end{tabular}
    \caption{
        Split statistics for bias analysis.
    }
    \label{tab:data_stat}
\end{table}

\subsubsection*{Additional analysis of generalization between populations}
\begin{figure}[htb]
    \centering
    \includegraphics[width=0.6\textwidth]{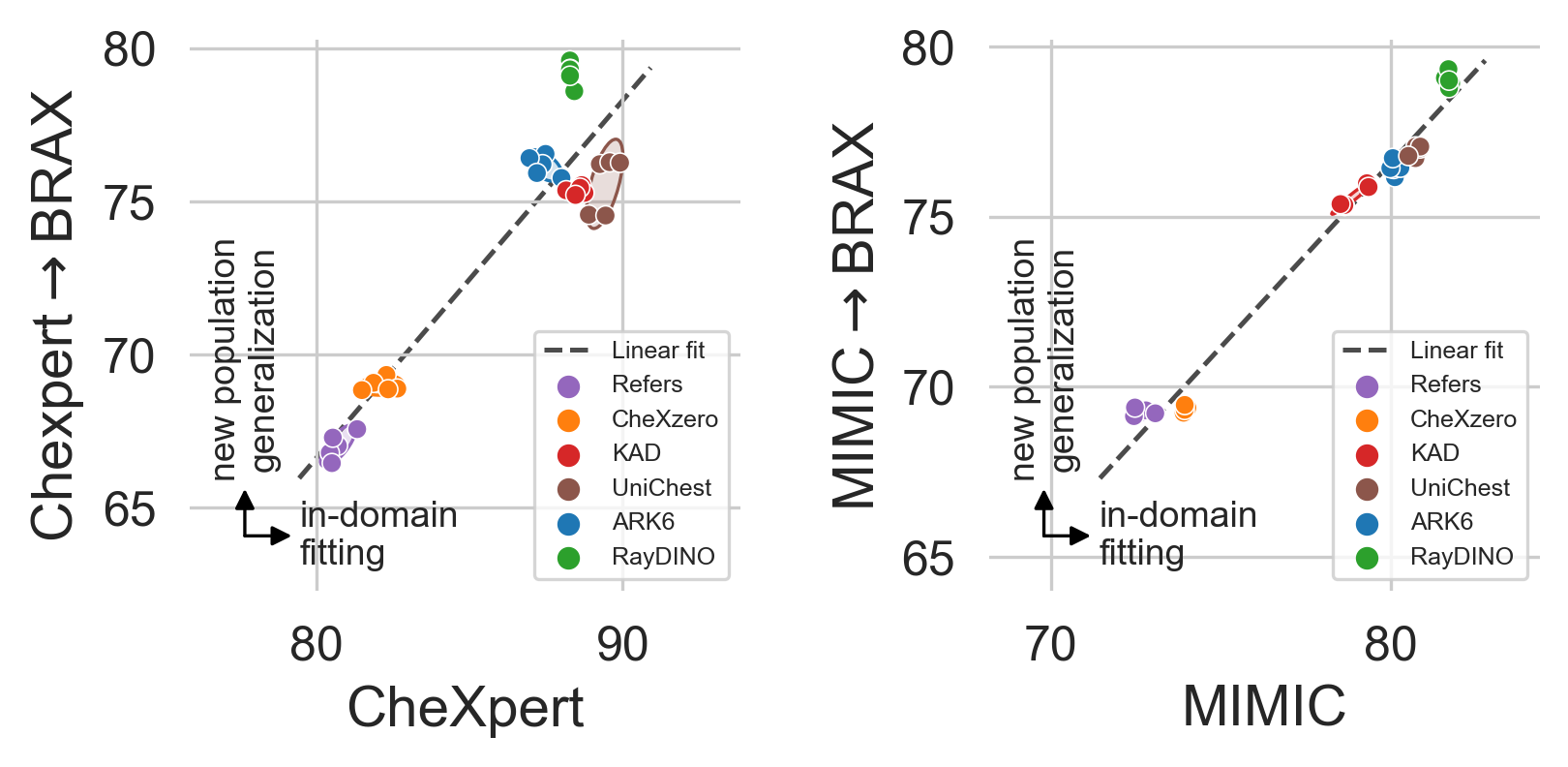}
    \caption{\textbf{In-Domain versus Out-Of-Domain performances:} We compare for each model the correlation between performances on the internal datasets CheXpert and MIMIC and the external dataset BRAX. Colors correspond to the backbone model and each of the 5 dots of the same color correspond to a different seed. We report the 95\%CI as ellipses.}
    \label{fig:scatter_transfer}
\end{figure}

We have shown that our model get excellent robustness when applied to different patient demographics.
The observation supporting this claim is the strong generalization of decoders trained on CheXpert or MIMIC when transferred to BRAX.
In fig.~\ref{fig:scatter_transfer}, we show the performance on the transfer dataset (BRAX) as a function of performance on the source dataset (CheXpert or MIMIC). 
We plot performance for our model and all competing backbones. Each dot represents one trial, and we draw the two-dimensional confidence interval.
We also represent the linear regression that we fit on these data points. 
Models above the plot's linear fit exhibit stronger generalization performance.
First, we see that RayDINO exhibits much better generalization, with a significant gap with the linear fit, especially when transferring from CheXpert.
Second, while our model achieves the best performance on BRAX, its performance on the source CheXpert data is below that of UniChest, indicating strong overfitting of the competing method on the CheXpert dataset.

It's also noteworthy that all competitor models have been trained in a supervised manner using MIMIC labels. When RayDINO's classifier is trained using labels from MIMIC instead of CheXpert but still tested on CheXpert, it achieves a performance of $89.4$ AUROC (95\%CI [$89.2$-$89.6$]), comparable to the best competitor ($-0.02$, $p=0.85$). This suggests that MIMIC may contain higher quality labels than CheXpert and that the comparison in Fig~\ref{fig:fig2}a and fig.~\ref{fig:scatter_transfer} against ARK6 and UniChest on CheXpert is skewed in their favor.

\subsubsection*{Generalization results from 2D CT-scan projection to real-world X-rays.} 
\begin{figure}[htb]
    \centering
    \includegraphics[width=\textwidth]{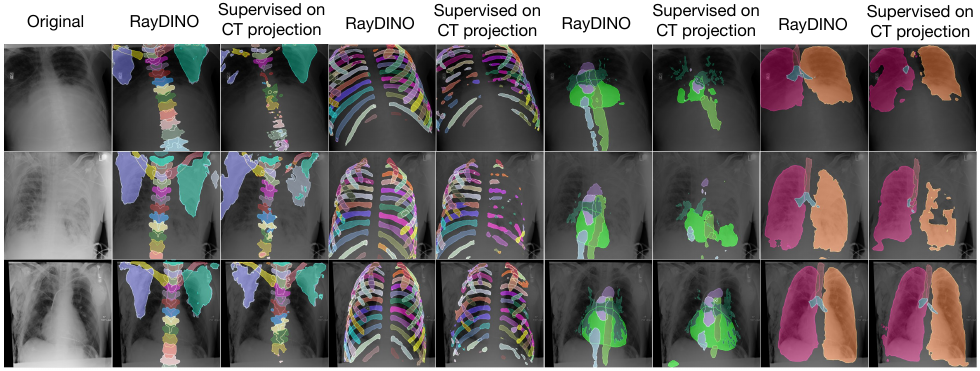}
    \caption{\textbf{Generalization of segmentation on real X-rays:} We present  examples of RayDINO's segmentation adapter trained on 2D CT-scan projection and evaluated on real-world X-rays. We compare it with a strong supervised baseline CXAS~\cite{Seibold_2023_CXAS}.}
    \label{fig:generalization_segmentation}
\end{figure}

To supplement the quantitative analysis presented in the Results section, we display in fig.~\ref{fig:generalization_segmentation} real-world X-rays from the NIH dataset. These images are segmented using RayDINO's adapter, trained on PAXRay++ and CXAS~\cite{Seibold_2023_CXAS}, a supervised segmentation model also trained on PAXRay++. RayDINO demonstrates the ability to handle more complex images with changes in patient's position and large opacities due to diseases, compared to the supervised model. This is achieved through the use of RayDINO's frozen features, which are trained on real-world X-rays, preventing overfitting on synthetic 2D projections of CT-scans and ensuring familiarity with real-world X-ray distributions.

RayDINO's ability to effectively transfer from 2D CT-scan projections to real chest X-rays is promising. This generalization could pave the way for labeling difficult diseases where practitioners annotate CT-scans as a gold standard and project the labels in 2D. RayDINO could then learn from these synthetic 2D data and transfer the knowledge to real-world X-rays. This approach could potentially reduce annotation errors and, ultimately, improve X-ray diagnostic performance while lowering the cost barrier for medical access.

\subsubsection*{Additional results for the explainability of RayDINO.}
\begin{figure}[htb]
    \centering
    \includegraphics[width=\textwidth]{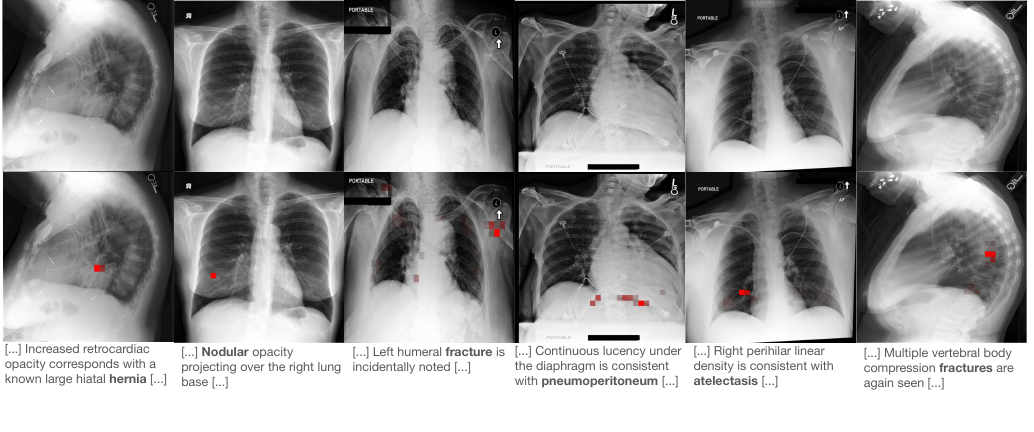}
    \caption{\textbf{Correlating attention maps with radiologist reports:} We present additional examples of RayDINO's attention maps alongside radiologist reports (specific finding of the classifier highlighted in bold) to demonstrate the accurate localization of the interpretability of RayDINO.}
    \label{fig:explainability_supp}
\end{figure}

\begin{figure}[htb]
    \centering
    \includegraphics[width=0.6\textwidth]{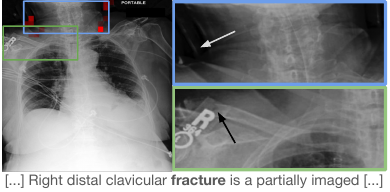}
    \caption{\textbf{Identifying failure case through RayDINO's interpretability:} We highlight a failure case where RayDINO incorrectly predicts a fracture by focusing on the patient's rigid neck brace (top rectangle in blue, white arrow). It fails to detect the actual clavicle fracture (bottom rectangle in green, black arrow) due to its partial imaging.}
    \label{fig:failure_case_supp}
\end{figure}

We show additional explainability images in fig.~\ref{fig:explainability_supp}. We can see that RayDINO can accurately pinpoint findings from all kind of categories on frontal and lateral X-rays. As an example of how RayDINO's explainability can be harnessed in clinical settings, we show in fig.~\ref{fig:failure_case_supp} a RayDINO's classification failure case where the model correctly predicts a fracture but using a wrong spurious correlation. We can see that RayDINO is looking in the top blue rectangle at a neck brace (white arrow) to predict a fracture and miss the badly imaged clavicle fracture in the lower green rectangle (black arrow). Neck braces probably appear more on patients with fractures and thus the model associated such information with the class `fracture'. Explainable artificial intelligence such as RayDINO can help radiologist and researchers find such spurious correlation and train better classifier using corrected training sets while making more informed analysis of X-rays.

\end{document}